# PLANNING, SCHEDULING, AND UNCERTAINTY IN THE SEQUENCE OF FUTURE EVENTS*

Barry R. Fox[1] and Karl G. Kempf[2]

*ABSTRACT: Scheduling in the factory setting is compounded by computational complexity and temporal uncertainty. Together, these two factors guarantee that the process of constructing an optimal schedule will be very costly and the chances of executing that schedule are very slight. Temporal uncertainty in the execution time environment can be offset by several methods: eliminate uncertainty by careful engineering, reduce uncertainty by using more powerful sensors, restore certainty whenever it is lost, quantify and circumscribe the remaining uncertainty. Unfortunately, these methods focus exclusively on the sources of uncertainty and fail to apply knowledge of the tasks which are to be scheduled. A complete solution must seek to adapt the schedule of activities to be performed according to the evolving state of the world. The example of vision directed assembly is presented which illustrates that the principle of least commitment, in the creation of a plan, in the representation of a schedule, and in the execution of a schedule, enables a robot to operate intelligently and efficiently, even in the presence of considerable uncertainty in the sequence of future events.*

[1]McDonnell Douglas Research Laboratories
Artificial Intelligence Technology Group
Dept 225/Bldg 305/Level 2E
POB 516, St. Louis, MO, USA 63166

[2]Artificial Intelligence Center
FMC Corporation
1185 Coleman Avenue
Santa Clara, CA, USA 95052

## 1. Statement of the Problem

The problem of sequencing and scheduling is to determine, in advance, when some set of events and activities should occur. The events and activities may be subject to some ordering constraints; the time of occurrence or the duration of some activities may be fixed; there may be some cost associated with the time that an activity is performed or with the rate with which it progresses. The times selected for these events and activities must conform to the constraints, expected times and durations, and must generally minimize the associated costs. In problems which are over constrained, the selected times must at least conform to a rational relaxation of the given constraints.

Scheduling in the factory setting is compounded by two factors[1]. The first is *complexity*. Most scheduling problems are NP-hard. The number of potential schedules grows exponentially with the number of activities to be scheduled and the number of resources to be utilized. It is simply impossible to itemize and evaluate every possible schedule in order to find the best schedule. At the same time, there is no known method for finding the best schedule which does not exhaustively consider (at least implicitly) every possible schedule. The second factor is *uncertainty*. Any schedule constructed in advance is based solely upon assumptions about the ordering, the times of occurrence and the durations of the activities.

The execution time environment rarely conforms to those assumptions. A schedule may be optimal with respect to some set of assumptions but it may be very costly to force the execution of the given tasks according to that schedule in the context of the actual execution times and durations.Together, these two factors guarantee that the process of constructing an optimal schedule will be very costly and the chances of actually executing that schedule are very slight.

A simple assembly task can be used to illustrate one aspect of temporal uncertainty. The task is to assemble a small gearbox consisting of the 10 parts shown in Figure 1. (For brevity the part names are reduced to just the first 2 letters in the remaining figures and tables.) Those parts are randomly packed into a small bin and delivered to a vision equipped robot which must acquire the parts from the bin and perform the assembly. This is identical to an assembly task which has been developed for the Air Force as a joint venture by SRI, Honeywell, and Adept Technologies[2], and is quite similar to an experimental assembly system studied at Edinburgh[3]. The characteristic feature of this task is that it is impossible to predict or control the order in which the parts will be recognized. Clearly, this can be best categorized as a Markov Process. Unfortunately, with even a few parts, it is impossible to itemize all of the possible states of the process and the relative probabilities of the state transitions. (Hence, this problem illustrates two kinds of uncertainty: uncertainty about the ordering of parts, and uncertainty about the statistical properties of the orderings.) The problem is to produce a working system which can perform this task efficiently, intelligently, and in a cost effective fashion.

## 2. Candidate Solutions

### 2.1 Eliminate Uncertainty

The most frequently proposed solution to this problem is to eliminate the uncertainty in the availability of the parts by re–engineering the task. Instead of delivering the parts in a bin, deliver the parts affixed to a pallet with specific locations and orientations. The robot can then be programmed to perform the task according to a fixed sequence of operations based upon the fixed locations of the parts. Of course the cost of this solution is determined by the cost of engineering the pallets and transport. The cost may be justifiable in large volume applications but it is unacceptable in the aerospace industry, for instance, where millions of parts are produced in quantities of 100 or less per year. The inventory of pallets and fixtures would exceed the value of the parts to be produced. Moreover, in many circumstances human labor would be required to prepare the pallets. The motive for introducing vision in such applications is to automate the production without introducing the added expense of custom fixtures, pallets, and transport.

### 2.2 Reduce Uncertainty

A second solution is to apply more powerful sensors in order to extract more information from the world. This is the approach employed by SRI, Honeywell, and Adept Technologies, in the project cited above. Their goal was to develop sophisticated vision hardware and software so that the robot could find exactly the part required at each step of the assembly. Of course such efforts might reduce the uncertainty in the process but they can never completely remove the uncertainty. Hence, the proposed system included an escape mechanism. If the required part could not be located the robot was programmed to shake the bin or stir the parts with the expectation that this might reveal the necessary part. Unfortunately, there is no way to predict how many shakes are necessary to reveal an obscured part and the execution time for the assembly will vary wildly[4]. The cost of this solution is the cost of the most sophisticated, and probably the most expensive vision system available, yet the benefit of this solution carries with it a high degree of uncertainty.

### 2.3 Restore Certainty

A third solution is to restore certainty whenever it is lost. For example, in the system developed at Edinburgh cited above, the parts were made available in a loose heap. The parts were separated from the heap one at a time and according to their identity moved to fixed locations in the workspace. After all of the parts were identified and laid out, the actual process of assembly proceeded according to a fixed sequence of operations. Although the parts would most certainly be acquired from the heap in a random order, the process of placing the parts in fixed locations restored the order necessary to perform the assembly in a fixed order. The cost of this solution is determined by the added execution time required to lay out the parts. Although the number of motions required to perform the assembly is fixed (one motion to acquire and buffer each part, and one motion to install each part) on the average, many of the intermediate motions are extraneous[4]. An intelligent solution would move a part to a buffer location *only if* it cannot be immediately installed in the assembly.

## 2.4 Quantify Uncertainty

Yet another solution might be to gather statistics over hundreds of trials and to program the robot to perform the assembly over the sequence of parts most likely to occur. When the sequence of parts deviates from the programmed order, the robot could resort to some other strategy such as shaking or buffering. This solution carries much of the same costs as the previous solutions and incurs the added cost of performing the experiments and gathering the statistics. Such costs are unjustifiable in small volume productions.

## 3. The Proposed Solution

It is unlikely that a single strategy or methodology is sufficient in a multifaceted problem such as this. Instead a good strategy would be to form a composite of the best available hardware and software technologies, incorporating each to the degree that it is useful and cost effective. Unfortunately there is a common weakness in the candidate solutions which must be addressed. The solutions discussed above focus exclusively on the process of acquiring the parts and they fail to apply knowledge of the assembly task itself. While there is no way to precisely predict or completely control the sequence of parts as they are acquired from the bin, there may be many feasible sequences for installing the parts in the assembly. The intelligent solution is to exploit this inherent flexibility and to *opportunistically schedule* the sequence of assembly operations at execution time, according to the availability of parts and according to the constraints on the assembly operations.

The implementation of an opportunistic scheduling strategy depends upon consistent application of the principle of least commitment. It is not unusual for a planner to determine the steps of a task and also a single fixed sequence for their execution. However, most tasks can be executed according to many different sequences. It is very unlikely that one, chosen before execution time, will be the best. It is certain, however, that if only one sequence is passed to the robot it will have no flexibility in the execution of that task. In order to maximize flexibility, a plan should consist of a set of steps and a *minimum* set of ordering constraints. Likewise, a schedule should consist of the sequence of steps that have been completed plus the set of *all possible* sequences for completing the remaining steps. Given such a schedule, the execution time sequencer can capitalize on opportunities presented in the execution environment; sidestepping difficulties with a given ordering by following a different but equally valid sequence.

These concepts, however, lead to a unique problem of representation. It has been argued in a previous paper[5] that a representation for plans and schedules based upon simple partial orders is inherently inadequate and that a state–space representation is not sufficiently compact. The principle of least commitment dictates that the plan for a task encompass every feasible strategy and that a schedule incorporate every admissable sequence of steps. This in turn dictates a representation which completely and compactly encodes those sequences. A sufficient representation, based upon sets of partial orders, is outlined in the paper cited above but a new style of planning, suitable for problems with a high degree of temporal uncertainty, remains to be explored.

Application of the principle of least commitment also influences the specific sequence of operations chosen at execution time. When several parts are recognized and accessible in the bin, and when more than one of those parts can be immediately included in the assembly the robot must pick one. The robot may consider the difficulty of acquiring a specific part, or the difficulty of installing that part, but it must give some preference to the operation which preserves the most options in sequencing the remainder of the assembly.

Consistent application of the principle of least commitment, in the creation of the plan, in the representation of the plan and schedule, and in the execution of the schedule, increases the likelihood that each time the robot consults the vision system, at least one of the parts currently visible in the bin can be immediately installed in the assembly. Given this flexibility, the robot can determine the actual sequence of operations at execution time, according to the sequence of parts acquired from the bin and according to the constraints on the assembly operations. Of course, there will be some circumstances when none of the visible parts can be immediately installed. In that case, one of the visible parts can be buffered thereby introducing some certainty into the process. As a whole, the strategy of opportunistically selecting parts to be installed or buffered enables the robot to operate efficiently, and intelligently in spite of the considerable uncertainty in the sequence parts acquired from the bin.

## 4. Empirical Results

Assume that the gearbox, shown in Figure 1, is to be assembled as described above. The parts are packed randomly into a small bin and delivered to a vision equipped robot which must acquire the parts from the bin, one at a time, and perform the assembly. Assume also that the robot will perform the assembly according to the strategy shown in Figure 2. (In fact, the plan for the assembly may consist of several distinct strategies, but for the sake of simplicity, only one strategy will be considered in this example.) Within that diagram each node denotes the operation of installing a part and each directed arc denotes an ordering constraint. For instance, the arc from **ri** to **co** denotes the constraint that the **ring gear** must be installed before the **cover**. Whenever multiple arcs converge at a single node, all of the predecessor steps must be completed before the operation denoted by the node can be executed.

Suppose that the robot has been programmed to perform the assembly according to the strategy of opportunistic scheduling. At each step of the assembly at least one part is visible and available in the bin and there may also be some parts available which have been previously moved from the bin to fixed layout locations. The robot will always install any part which is presently available and can be immediately included in the assembly. If none of the available parts can be immediately installed it will choose one of the parts in the bin and move it to a fixed location for later use. If more than one part can be installed the robot should choose the one which preserves the most future sequencing options, and if none can be installed the robot should choose to buffer the part which will be needed earliest in the assembly. The sequence of operations shown in Table 1 might be typical of the course of action generated by an opportunistic scheduling strategy.

According to the simulation studies reported in an earlier paper[4], this technique is far superior to the individual methods of buffering or shaking discussed above.

## 5. Conclusion

Several different techniques can be applied in problems which involve task execution in the presence of temporal uncertainty. Wherever possible, uncertainty should be eliminated or reduced, certainty should be restored whenever it is lost, the remaining uncertainty should be quantified and circumscribed, but an invaluable technique is to exploit the inherent flexibility in the tasks to be performed and to implement a strategy of opportunistic scheduling. This requires consistent application of the principle of least commitment in the creation of the plan, in the representation of the plan and schedule, and in the execution of the schedule. A hybrid of these techniques enables a robot to operate intelligently, efficiently, and in a cost effective manner, in the presence of considerable uncertainty in the sequence of events.

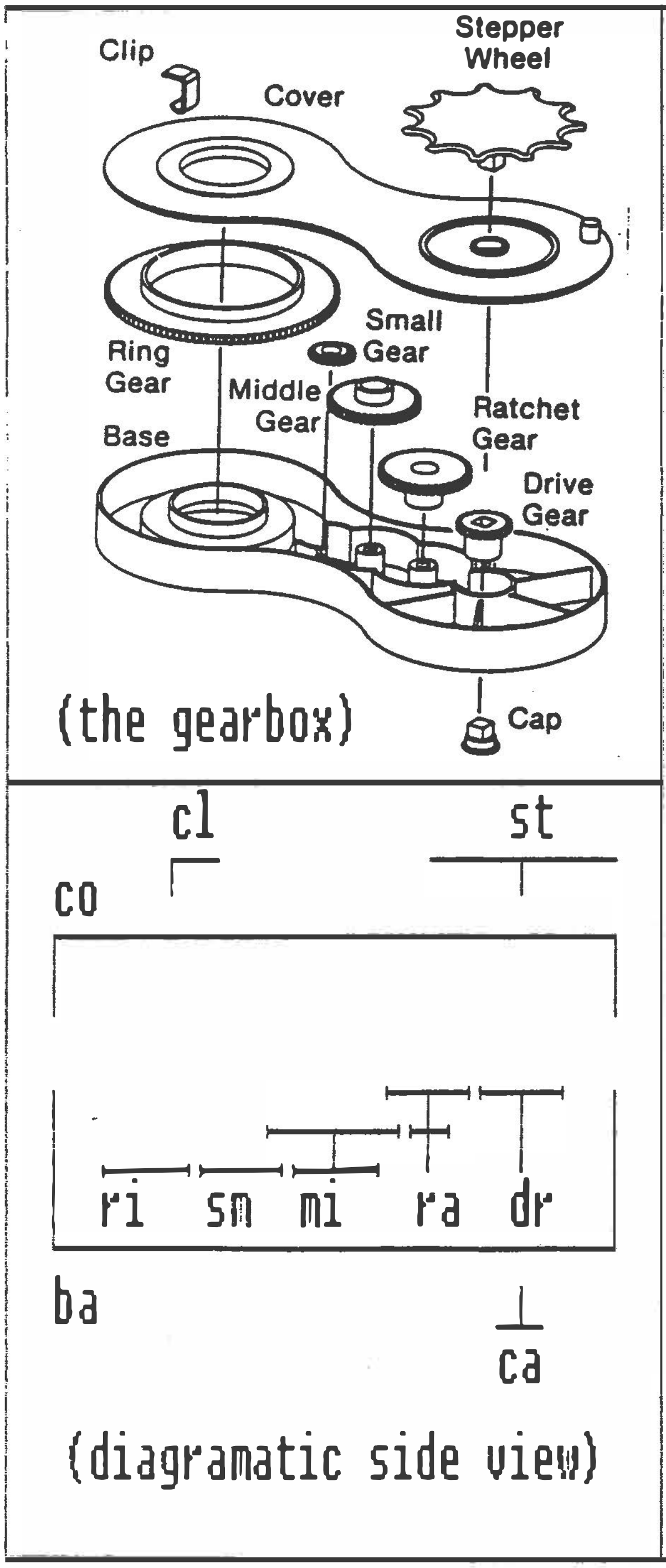

(the gearbox)

(diagramatic side view)

Figure 1

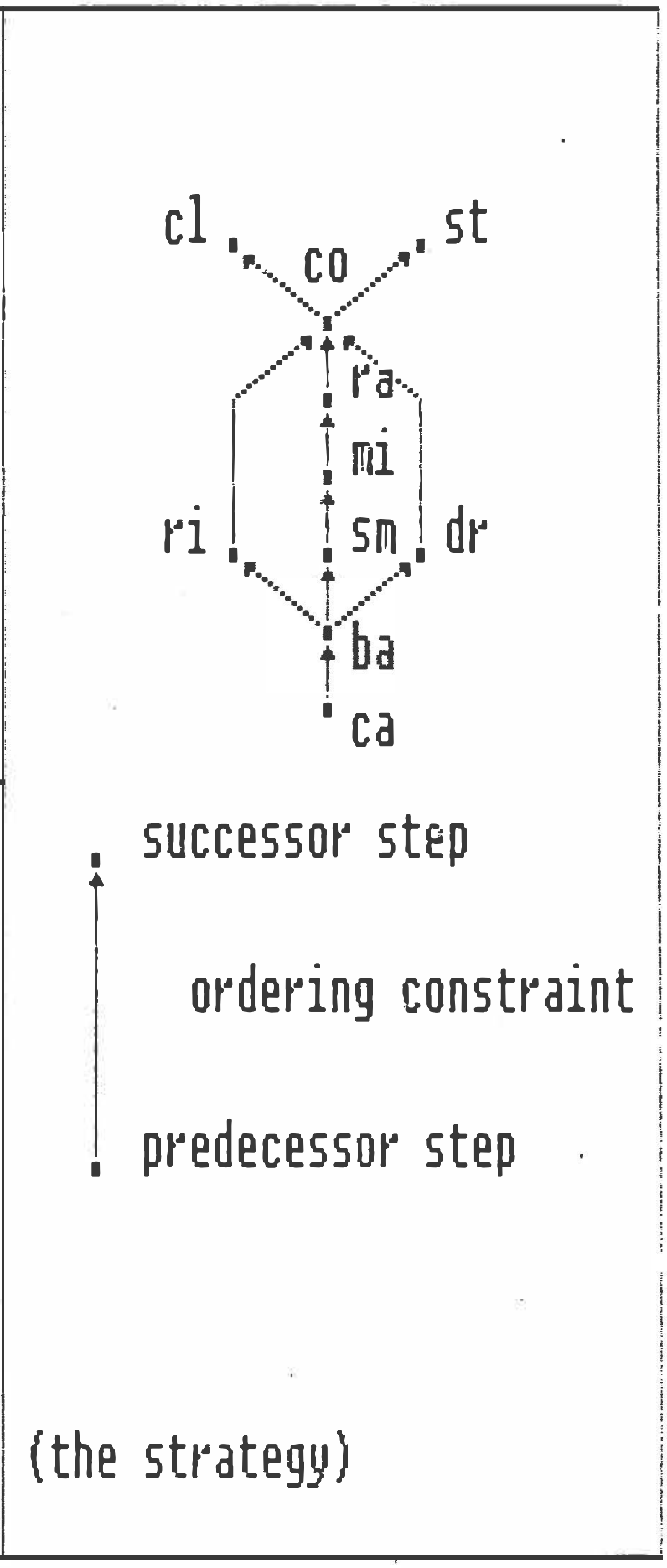

(the strategy)

Figure 2

| Visible | Buffered | Installed | Action |
|---|---|---|---|
| ri,ra | | | buffer ri |
| ca,co | ri | | install ca |
| co,mi | ri | ca | buffer mi |
| st,dr | ri,mi | ca | buffer dr |
| cl,ba | ri,mi,dr | ca | install ba |
| cl,co | ri,mi,dr | ca,ba | install ri |
| cl,co | mi,dr | ca,ba,ri | install dr |
| cl,co | mi | ca,ba,ri,dr | buffer co |
| cl,sm | mi,co | ca,ba,ri,dr | install sm |
| cl,st | mi,co | ca,ba,ri,dr,sm | install mi |
| cl,st | co | ca,ba,ri,dr,sm,mi | buffer cl |
| ra,st | co,cl | ca,ba,ri,dr,sm,mi | install ra |
| st | co,cl | ca,ba,ri,dr,sm,mi,ra | install co |
| st | cl | ca,ba,ri,dr,sm,mi,ra,co | install st |
| | cl | ca,ba,ri,dr,sm,mi,ra,co,st | install cl |
| | | ca,,ba,ri,dr,sm,mi,ra,co,st,cl | |

Table 1